\documentclass[10pt,twocolumn,letterpaper]{article}

\usepackage[pagenumbers]{cvpr} 
\makeatletter
\@namedef{ver@everyshi.sty}{}
\makeatother

\usepackage{booktabs}
\usepackage{times}
\usepackage{epsfig}
\usepackage{graphicx}
\usepackage{amsmath,xfrac,wrapfig,comment}
\usepackage{amssymb,amsthm}
\usepackage{mathtools}

\usepackage{dsfont}
\usepackage{booktabs,paralist,hhline,colortbl}
\usepackage{algorithm}
\usepackage{algorithmicx}
\usepackage{algpseudocode}

\algnewcommand\algorithmicinput{\textbf{Input:}}
\algnewcommand\algorithmicoutput{\textbf{Output:}}
\algnewcommand\Input{\item[\algorithmicinput]}%
\algnewcommand\Output{\item[\algorithmicoutput]}%
\newcommand{\algrule}[1][.2pt]{\par\vskip.5\baselineskip\hrule height #1\par\vskip.5\baselineskip}

\usepackage{graphicx}

\newtheorem{remark}{Remark}

\newcommand{\xx}{\ensuremath{\mathbf{x}}}

\usepackage{booktabs,paralist,hhline,colortbl} 

\definecolor{Maroon}{RGB}{204, 102, 0}

\definecolor{rulecolor}{rgb}{0.0, 0.06, 0.54}
\definecolor{tableheadcolor}{rgb}{0.74, 0.83, 0.9}
\definecolor{bluecolor}{rgb}{0.74, 0.83, 0.9}

\definecolor{imgbluecolor}{rgb}{0.28, 0.62, 0.97} 
\definecolor{imgorangecolor}{rgb}{0.89, 0.47, 0.18}

\usepackage[export]{adjustbox} 

\usepackage{tcolorbox}
\newtcolorbox{mybox}[3][]
{
  colframe = #2!15,
  colback  = #2!10,
  coltitle = #2!10!black,  
  title    = {#3},
  boxsep   = 0.25pt,
  left     = 0.5pt,
  right    = 0.5pt,
  top      = 0pt,
  bottom   = 0pt,
  width=\linewidth,
  #1,
}

\usepackage[pagebackref=true,breaklinks=true,letterpaper=true,colorlinks,bookmarks=false]{hyperref}

\usepackage[capitalize]{cleveref}
\crefname{section}{Sec.}{Secs.}
\Crefname{section}{Section}{Sections}
\Crefname{table}{Table}{Tables}
\crefname{table}{Tab.}{Tabs.}

\def\cvprPaperID{20} 
\def\confName{AICC workshop}
\def\confYear{2022}

\begin{document}

\title{Image2Gif: Generating Continuous Realistic Animations  with Warping NODEs}
\author{Jurijs Nazarovs\\
University of Wisconsin - Madison\\
{\tt\small nazarovs@wisc.edu}
\and
Zhichun Huang\\
Carnegie Mellon University\\
{\tt\small zhichunh@cs.cmu.edu}
\and
{
\mbox{\url{https://github.com/JurijsNazarovs/warping\_node}}
}
}

\maketitle

\begin{abstract}

Generating smooth animations from a limited number of sequential observations has a number of applications in vision. For example, it can be used to increase number of frames per second, or generating a new trajectory only based on first and last frames, e.g. a motion of face emotions. 
Despite the discrete observed data (frames), the problem of generating a new trajectory is a continues problem. In addition, to be perceptually realistic, the domain of an image should not alter drastically through the trajectory of changes.
In this paper, we propose a new framework, Warping Neural ODE, for generating a smooth animation (video frame interpolation) in a continuous manner, given two (``farther apart'') frames, denoting the start and the end of the animation.
The key feature of our framework is utilizing the continuous spatial transformation of the image based on the vector field, derived from a system of differential equations.  This allows us to achieve  the smoothness and the realism of an animation with infinitely small time steps between the frames.
We show the application of our work in generating an animation given two frames, in different training settings, including Generative Adversarial Network (GAN) and with $L_2$ loss.

\end{abstract}

\vspace{5pt}
\section{Introduction}
\label{sec:intro}
Conventional video imaging sensors can only capture a moving scene with limited frame rates, a constraint that hinders it from vividly recording the highly dynamic and volatile physical world. Video Frame Interpolation (VFI) is a technique that address this issue by generating high-frame-rate videos using the rich visual information retrieved from low-frame-rate sources. The task is to synthesize the unavailable intermediate states in-between two or more consecutive video frames that together form a geometrically and temporally coherent sequence. VFI has wide applications in the industry, e.g., generating slow-mo videos \cite{jiang2018super}, efficient use of communication bandwidth, and potentially reducing the memory cost for storing large videos.

\begin{figure}[t]
    \centering

     \includegraphics[width=\columnwidth, trim={2cm, 0.2cm, 3.5cm, 0.2cm}, clip]{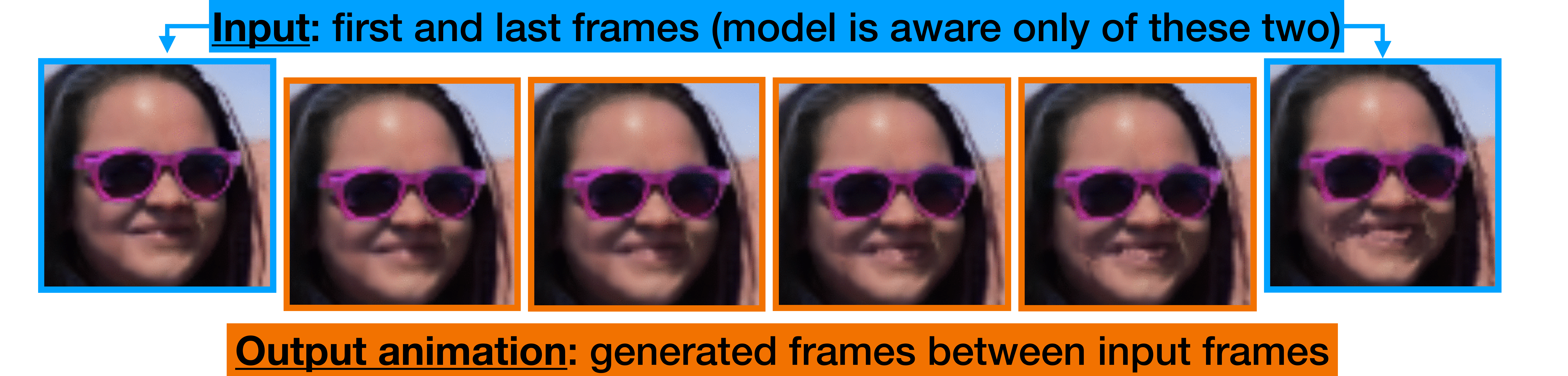}

    \caption{\footnotesize Our model is aware of only the first and last images (`blue' frames). No ground truth is available, the remaining imags are model predictions. Note that for visualization purposes we only show a few points along the animation trajectory:  a `.gif' is on the website.}
    \vspace{-5pt}
    \label{fig:teaser}
\end{figure}

A challenging setting in VFI is to create a short animation based on two frames that could in principle be conceptually farther apart. 
This makes it harder to interpolate compare to a typical `increase FPS' task with very similar nearby frames.
Consider an example in Figure~\ref{fig:teaser}, where we may be given two images of a person: non-smiling and smiling (first and last blue frames). Generating the full path of smiling emotion, orange frames, creates an ``alive'' effect from a no-smile to smile (similar to live-photos). 
The challenge lies in the sizable difference between the no-smile and smile frames.

{\bf Contributions.} 
In this work, we propose a new method for VFI, Warping Neural ODE, to generate an animation based entirely on the two input images (or video frames), thus assuming no additional information is available, e.g. from an event camera \cite{mueggler2017event}.
Our method models the transformation between frames with a sequence of diffeomorphisms according to a system of differential equations \cite{chen2018neural}. This allows smooth forward and backward warping among the generated frames, thus implicitly enforcing the spatial and temporal cohesiveness. 
Compared to existing approaches based on deep generative networks \cite{karras2019style}, our method does not suffer from hallucination (i.e. falsely producing non-existent visual elements) and is able to interpolate between frames at arbitrary temporal resolutions (infinitely small time steps between frames) due to properties afforded directly due to the use of diffeomorphisms. 

\section{Method}
\begin{figure}[t]
    \centering
    \includegraphics[width=\columnwidth, trim={0cm, 2cm, 1cm, 0cm}, clip]{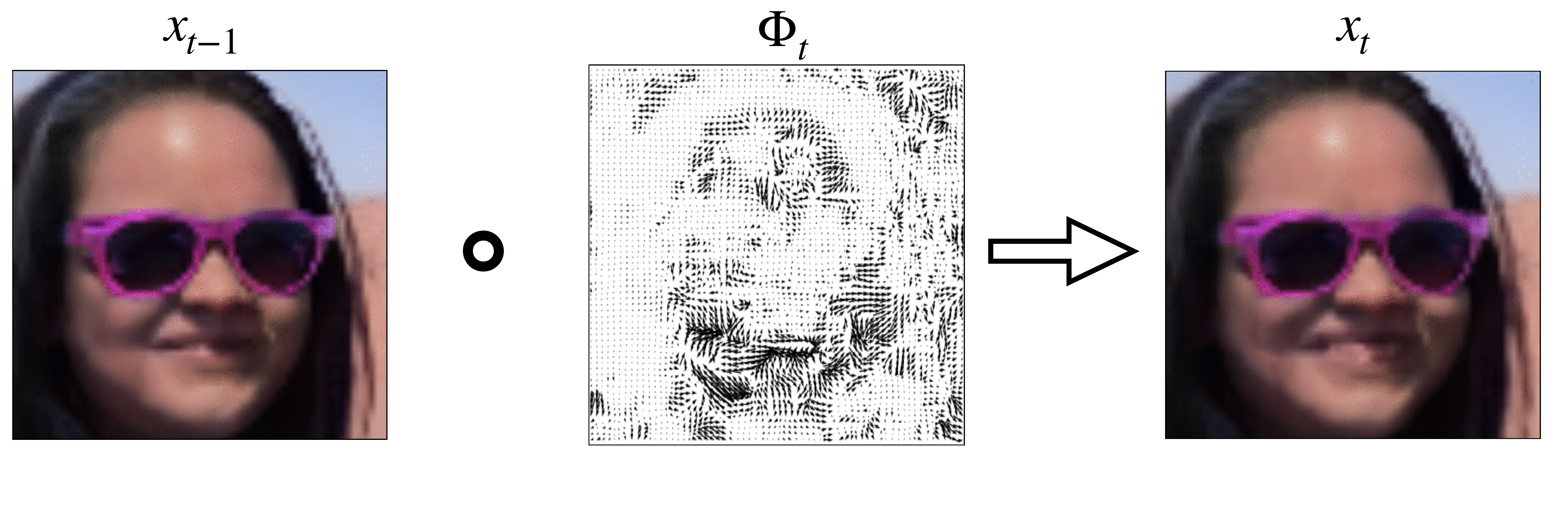}
    
    \caption{\footnotesize
    Given a vector field $\Phi_t$ we can warp image $x_{t-1}$ to $x_t$ in a natural for an eye way, without making drastic changes between frames.
    }
    \vspace{-15pt}
    \label{fig:model}
\end{figure}

To generate an animation, we need a collection of images $\xx = \{x_0, \ldots, x_t, \ldots, x_1\}$, where we define $x_0$ as a first frame, $x_1$ as a last frame, and $x_t$ as a frame at any time step $t\in(0, 1).$
Given a fixed amount of time for which the animation is played, the `smoothness' of the animation $\xx$ is defined by the size $|\xx|$. 
To increase the smoothness of an animation and make it perceptually natural, there are two desired requirements:
\textit{First}, we need to be able to generate $x_t$ at any time step $t\in (0, 1)$ with an infinitely small difference between steps $\Delta t_i = t_{i+1} - t_i$, i.e., $\Delta t_i \rightarrow 0$.
\textit{Second}, the changes between frames $x_{t_i}$ and $x_{t_{i+1}}$ should be carried in a minimal way, and ideally without introducing new information with respect to colors contained in the image, see Figure~\ref{fig:model}.

We start the motivation for this work with the second requirement, namely, colors, which have not previously 
been seen in the image at time step $t-1$, should not appear at $t$. This makes the animation look natural. For this reason, we seek to find a spatial diffeomorphism warping $\Phi_t$, to transform an image $x_{t-1}$ to $x_t$.
The notion of diffeomorphism is important since it guarantees the invertibility of the transformations, which conserves topological features \cite{rousseau2020residual}. For us, it means avoiding bringing/loosing color information during warping.
Next, given a warping $\Phi_t$, we should be able to define a warped image at any time step $t$.
Then the technical question is, how can we define $\Phi_t$ to be a diffeomorphism warping at any time step $t\in (0, 1)$, such that $\Delta t_i \rightarrow 0$? The rest of the section is motivated by our recent work \cite{nazarovs2022rf}.

\subsection{Learning diffeomorphism warping}
In general, a warping is considered to be a spatial transformation defined on the domain of the images. In other words, when we apply a warping $\Phi$ on an image $x$, we actually transform the coordinate system $S$ of an image $x(S)$ \cite{jaderberg2015spatial}. There are different ways to transform a coordinate system. For example, it can be carried out either through the definition of: (a) rotation matrix or (b) vector field, which applies to a pixel domain $S$.
A vector field warping is considered to be more flexible, since it allows each pixel of $S$ to move in an independent direction. For this reasons, we define warping $\Phi$ as a vector field, more on this in Remark~\ref{rem:vf}.

\begin{figure}[t]
    \centering
    \includegraphics[trim={0cm, 0cm, 2cm, 0cm}, clip, width=\columnwidth]{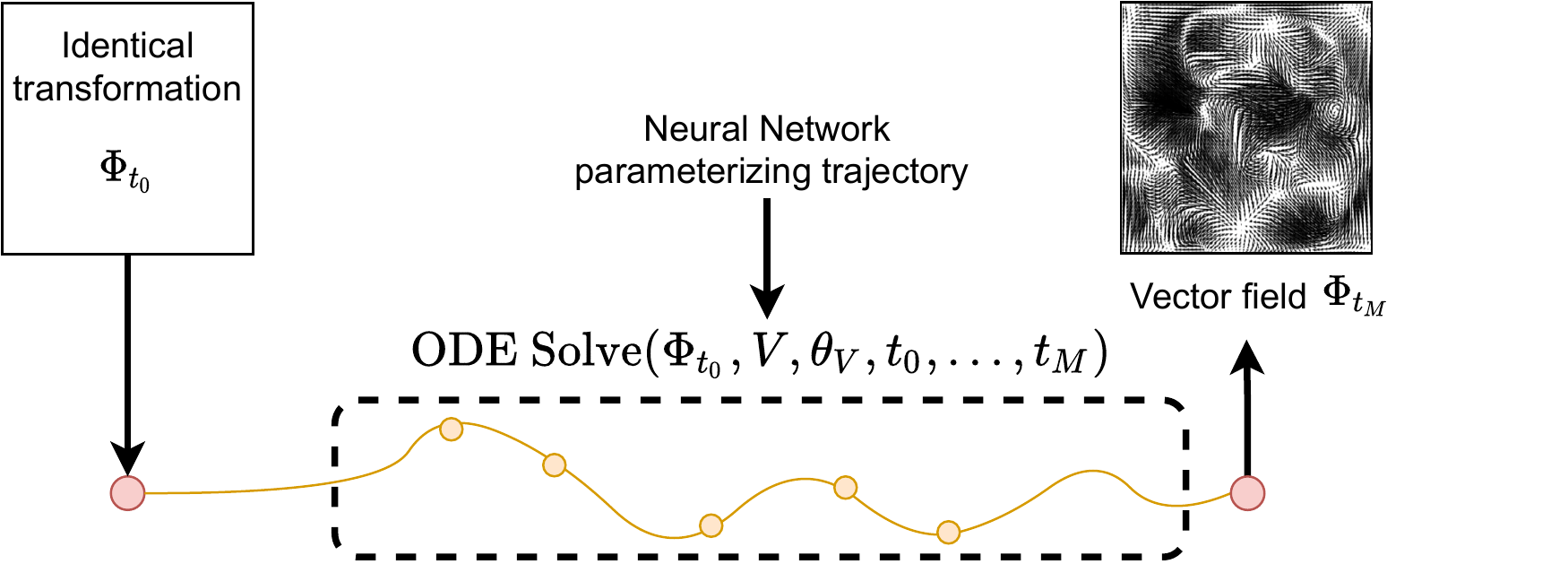}
    
    \caption{\footnotesize
    Warping Neural ODE models diffeomorphism $\Phi$ as a solution of ODE \eqref{eq:diff}, where RHS is modelled by NN. Resulted transformation $\Phi_{t_M}$ is applied to the coordinate system of the image to generate warped domain.
    }
    \vspace{-15pt}
    \label{fig:warpode}
\end{figure}

A specific class of diffeomorphism spatial operations, which define a subgroup structure in the underlying Lie group\cite{iserles2000lie}, can be parameterized by an ordinary differential equation (ODE)\cite{rousseau2020residual,ashburner2007fast}:
\begin{equation}
\frac{\mathrm{d} \Phi_t}{\mathrm{d} t}=V(\Phi_t),
\label{eq:diff}
\end{equation}
where $\Phi_t$ is the diffeomorphism at time $t$, and $V$ the stationary velocity vector field. 
{\it Forward warping:} by starting from the initial point (identity transformation) $\Phi_0$, we are able to integrate \eqref{eq:diff} in time ($t:0\rightarrow 1$) to obtain  $\Phi_1$, such that $x_1(S) = x_0(\Phi_1(S))$ and of course, generate $x_t = x_0(\Phi_t(S))$. 
While the diffeomorphism warping was necessary for us to make an animation looks more natural, by avoiding introducing/losing colors between frames, another benefit is an ability to define a reverse animation, which can be useful in application setup, if only the final frame $x_1$ is available.
{\it Backward warping:} in general with learning warping transformations, integrating backward in time ($t:1\rightarrow 0$) does not result in a reverse warping \cite{ashburner2007fast}. However, since \eqref{eq:diff} corresponds to a Lie group, it provides a definition of the exponential operators, the proper way to define backward warping $\Phi_{-1}$ is by integrating \eqref{eq:diff} over time ($t:0\rightarrow -1$).
To account for the richness of transformations, we parameterize the velocity $V$ as a network, which leads to {\bf Warping Neural ODE}, Fig.~\ref{fig:warpode}.

\subsection{Mechanisms for generating an animation}
Without loss of generality, consider that for generating the animation, we only observe two time steps of the whole animation `trajectory': the beginning $x_0$ and the end $x_1$, and our goal is to generate $x_t$ for $t\in(0,1)$.
Since $\Phi_t$ is defined as a solution of ODE in equation \eqref{eq:diff}, during training, the loss contains only information about $x_0$, $\Phi_1$ and $x_1$. Namely, given the warping $\Phi_1$, we need to make sure that $x_0(\Phi_1(s))$ is `equal' to $x_1$.
Depending on the available data, we can achieve it in different ways. 
For example, if data contains paired images, that is $x_0$ and $x_1$ are first and last frames of the same object (e.g. same smiling person), then the $L_2$ loss can be used for training to require that $x_0(\Phi_1(s)) = x_1$.
In contrast, if we do not have access to paired images of the animation, e.g. $x_0$ is non-smiling person $A$ and $x_1$ is a smiling person $B$, then there are different divergences, which can be used for training to achieve that $x_0(\Phi_1(s)) \sim x_1$, e.g. Jensen-Shannon \cite{fuglede2004jensen}, KL \cite{hershey2007approximating}, or simply minimize the Wasserstein (EM) distance \cite{ruschendorf1985wasserstein} between distribution of warped images $x_0(\Phi_1(s))$ and $x_1$ (our choice):
\begin{equation}
W\left(\mathbb{P}_{r}, \mathbb{P}_{g}\right)=\inf _{\gamma \in \Pi\left(\mathbb{P}_{r}, \mathbb{P}_{g}\right)} \mathbb{E}_{(x, y) \sim \gamma}[\|x-y\|],
\label{eq:wasser}
\end{equation}
where $\Pi\left(\mathbb{P}_{r}, \mathbb{P}_{g}\right)$ denotes the set of all joint distributions $\gamma(x, y)$ whose marginals are respectively $\mathbb{P}_{r}$ and $\mathbb{P}_{g}$. 
To achieve this, we minimize the efficient approximation of the Wasserstein distance similar to \cite{arjovsky2017wasserstein,gulrajani2017improved}. However, in contrast to GAN, in our setup the generator (Neural ODE component) does not generate images based on random samples, but is only used to create a warping $\Phi$ with no randomness.

\subsection{Final loss and Method summary} 

\begin{mybox}{gray}{}
\begin{center}
\begin{algorithm}[H] 
\caption{Learning diffeomorphism $\Phi: x_0 \rightarrow x_1$}
\label{algo:method}
\begin{algorithmic}[1]
    \footnotesize
    \Input Initial frame $x_0$ and final frame $x_1$
    \Output{Warping $\Phi_t$ $\implies$ animation $\xx=\{x_t\}_{t\in[0,1]}$}
    \algrule
    \Require  parameterized by Neural Networks:  $V$ in \eqref{eq:diff}. In GAN setup critic $D$ with $n_D$ number of critic updates
    \algrule
    \While{V has not converged}
        \State Set  $\Phi_0$ as identical transformation (vector field). 
        \State Using Neural ODE($V,\Phi_0)$ find a solution $\Phi_1$.
        \State Given $\Phi_1$, warp $x_0$ to $\hat{x}_1$
        \State {\bf If} paired images are available, compute $L_2(\hat{x}_1, x_1)$; {\bf Else} Run {\bf MinWasDist($\hat{x}_1$, $x_1$)} to minimize the Wasserstrein distance
    \EndWhile
    \Procedure{MinWasDist}{$\hat{x}_1$, $x_1$}
        \For{$i=0,\ldots, n_D$}
        \State update $D$ by minimizing {\it Critic's loss}:
        
        ~~~~$-D(Z) +D(\hat{x}_1) + \lambda GP(D)$
        \Statex \Comment GP is gradient penalty for Critic D \cite{gulrajani2017improved}~~~~~~~~
        \EndFor
        \State update $V$ by minimizing {\it ODE loss}:
        
        $-D(\widehat{x}_1)$ + JD + OG \Comment JD, OG  defined in Remark~\ref{rem:loss}
    \EndProcedure
\end{algorithmic}
\end{algorithm}
\end{center}
\end{mybox}

\begin{figure*}[!ht]
    \centering
    \begin{subfigure}{0.6\textwidth}
    \includegraphics[width=1\textwidth]{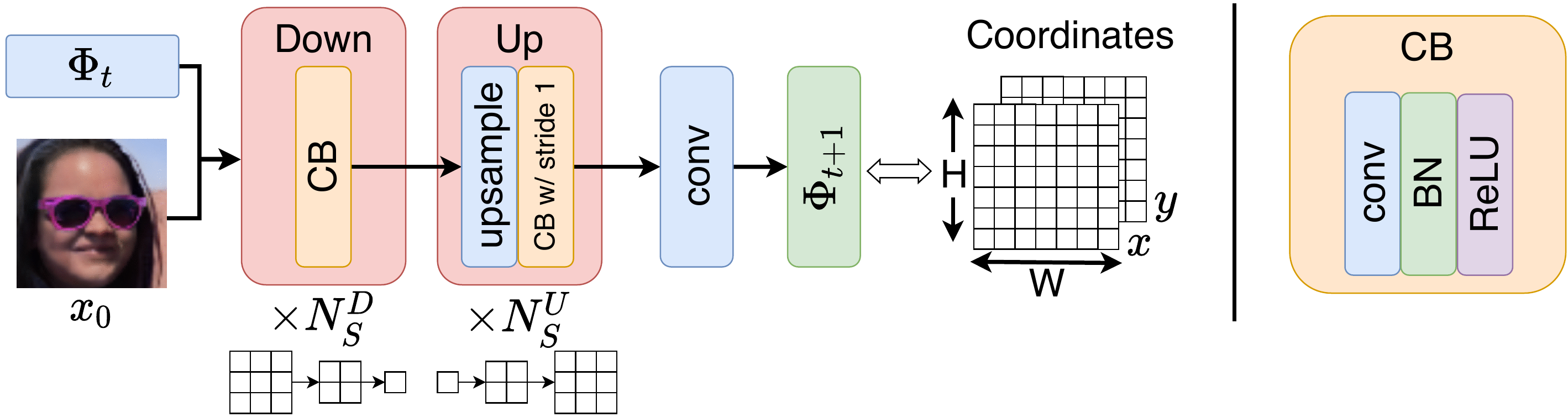}
    \caption{    }
    \end{subfigure}
    {~~~~ \vrule height 2cm width 1pt}
    \begin{subfigure}{0.3\textwidth}
    \includegraphics[width=1\textwidth]{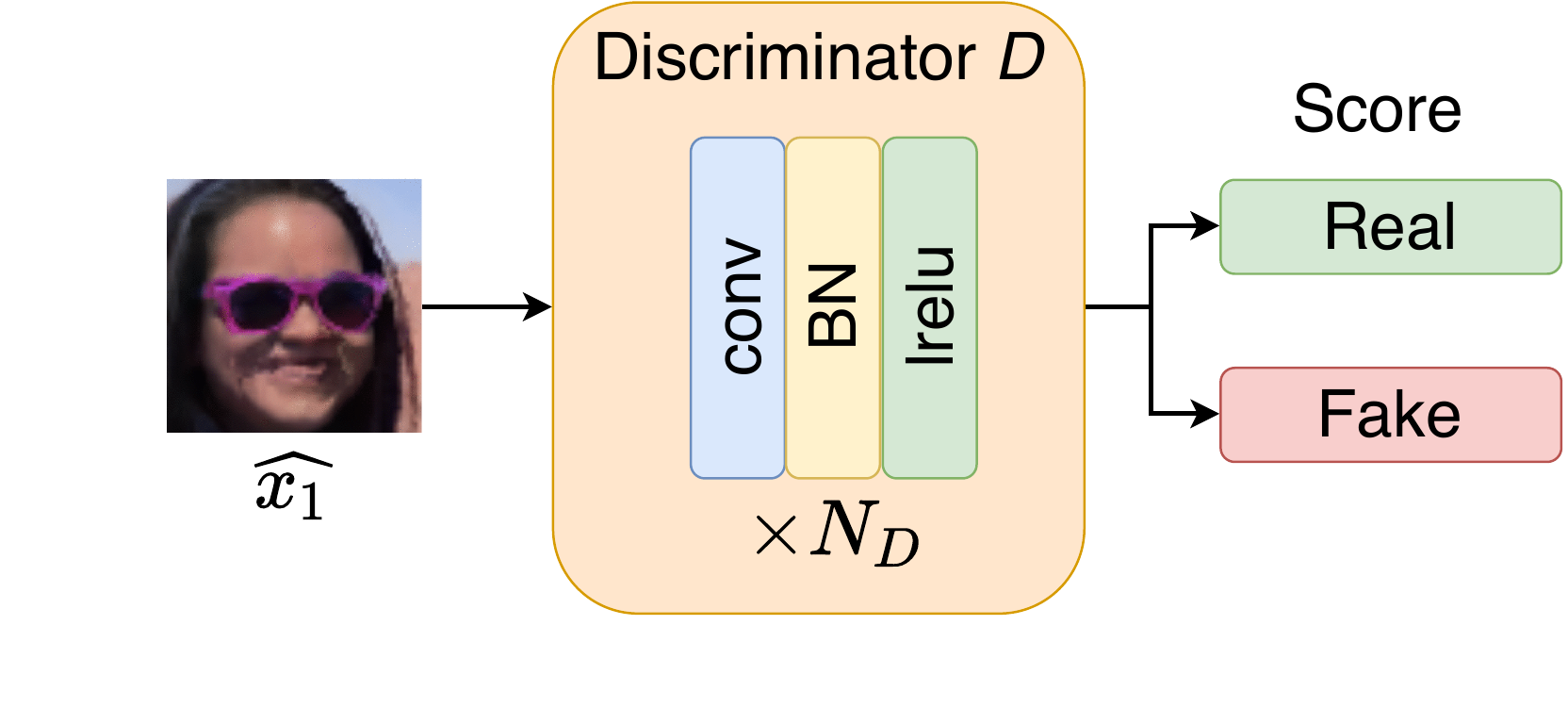}
    \caption{    }
    \end{subfigure}
    
    \caption{\footnotesize
     \textit{(a):} Warping Neural ODE part, which is used to generate the warping $\Phi_{t+1}$ at time point $t+1$.
     \textit{(b):} Discriminator $D$, main goal of which is to differentiate between warped image $\hat{x}_1$ and observed image $x_1$. It is used in case when we do not have paired images to generate animation. $N$ stands for number of convolution blocks.}
    \label{fig:gendisc}
    \vspace{-10pt}
\end{figure*}

\begin{remark}
\label{rem:vf}
As discussed earlier, to model warping $\Phi_t$ we follow a common technique \cite{jaderberg2015spatial,ashburner2007fast}. The main idea is to model warping as target coordinates, from which we sample (or move to). That is, for input image with dimension $h\times w$, $\Phi_t$ is of dimension $h \times w \times 2$, where 2 appears because the dimension of image is 2. Since we model warping $\Phi_t$ as NODE, the modelled derivative $V(\Phi_t)$ has to be of the same size $h \times w \times 2$. Since it is important to generate different warpings according to the input $x$, we model $V(\Phi_t)$ conditioning on image $x$, as $V(\Phi_T, x)$. We accomplish this using U-Net, demonstrated in Figure~\ref{fig:gendisc} (a) with number of layers/channels depending on a problem.  For the case where paired images are unavailable, we use the following discriminator  to minimize Wasserstrein distance, presented in Figure~\ref{fig:gendisc} (b).
\end{remark}

\begin{remark}
While theoretically, it is guaranteed that there is a unique solution to the system \eqref{eq:diff} given $\Phi_0$, see \cite{ohrnell2020lie} (pp. 8), to accelerate convergence, we add additional constraints (penalties) to the ODE loss in Algorithm~\ref{algo:method}, JD and OG respectively. Namely, we require (a) the Jacobian Determinant of each $\Phi_t$ to be non-negative \cite{kuang2019cycle}, to avoid collapsing several pixels into one, and (b) prevent generating warping $\Phi_t$, with vectors going outside the grid (image frame).
\label{rem:loss}
\end{remark}
\vspace{-20pt}
\begin{align*}
    JD &= \sum_t \sum_s\biggl( |JD(\Phi_t(s))| - JD(\Phi_t(s))\biggr) \\
    OG &=  \sum_t\sum_s\biggl( (\text{grid}(s) +\Phi_t - x_\text{size}) + (\text{grid}(s) +\Phi_t)\biggr)
\end{align*}
We note that OG computation is related to our implementation of warping, 
common in vision \cite{jaderberg2015spatial,ashburner2007fast}. We model $\Phi_t$ as a vector field, and consider the `grid' as a mesh coordinate system from 0 to size of image $x_\text{size}$. Then, pixels in `grid(s)' will be sampled from (or move to) location $\text{grid}(s) + \Phi_t$. The OG term prevents learning vector fields $\Phi_t$, which map outside the grid.

\begin{figure}[!ht]

     \includegraphics[width=0.159\columnwidth, trim={5cm, 3cm, 7cm, 4cm}, clip]{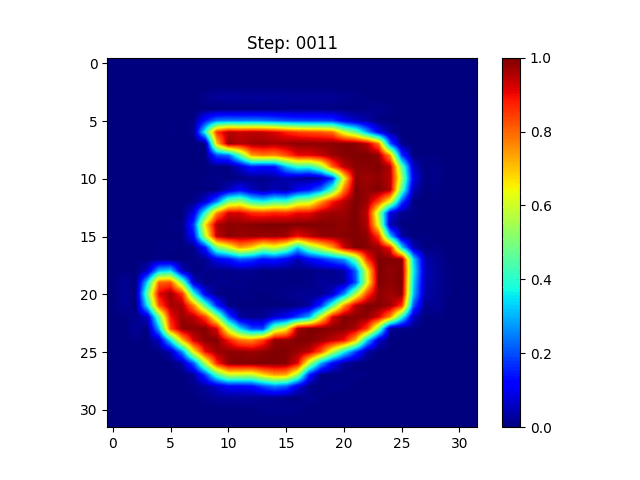}
     \includegraphics[width=0.159\columnwidth, trim={5cm, 3cm, 7cm, 4cm}, clip]{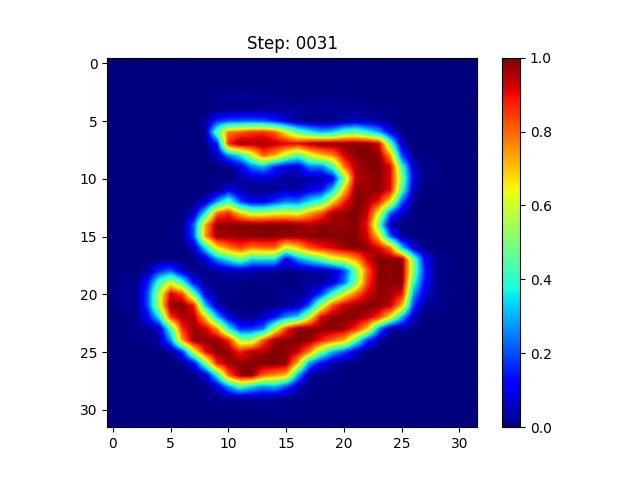}
     \includegraphics[width=0.159\columnwidth, trim={5cm, 3cm, 7cm, 4cm}, clip]{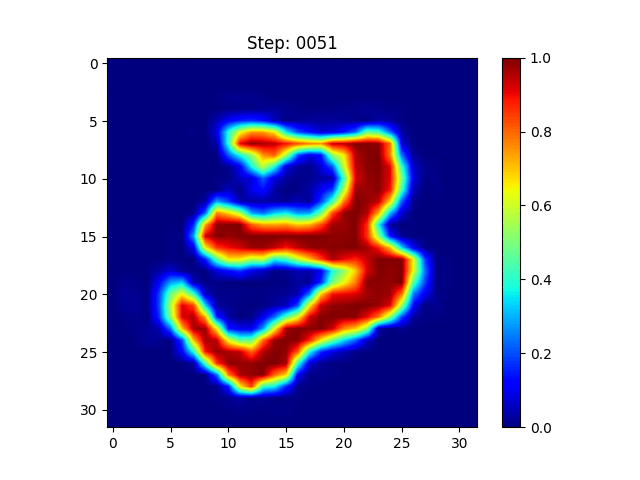}
     \includegraphics[width=0.159\columnwidth, trim={5cm, 3cm, 7cm, 4cm}, clip]{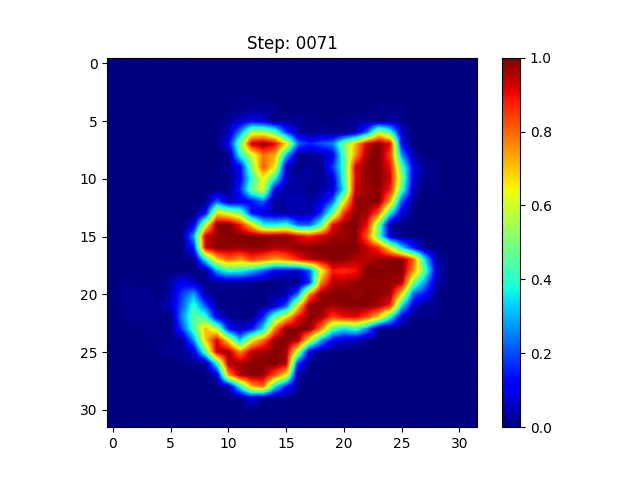}
     \includegraphics[width=0.159\columnwidth, trim={5cm, 3cm, 7cm, 4cm}, clip]{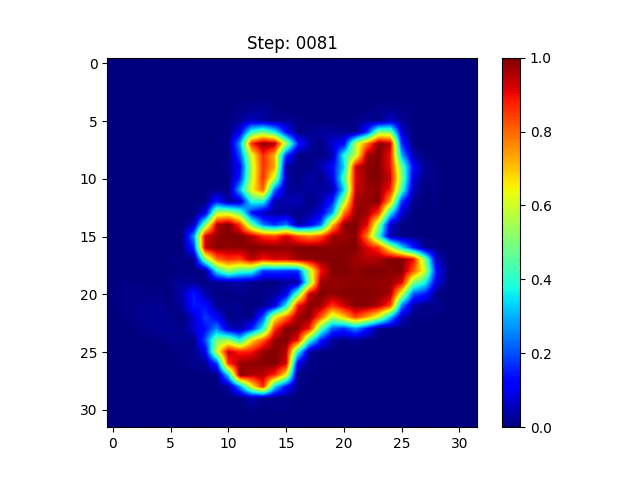}
     \includegraphics[width=0.159\columnwidth, trim={5cm, 3cm, 7cm, 4cm}, clip]{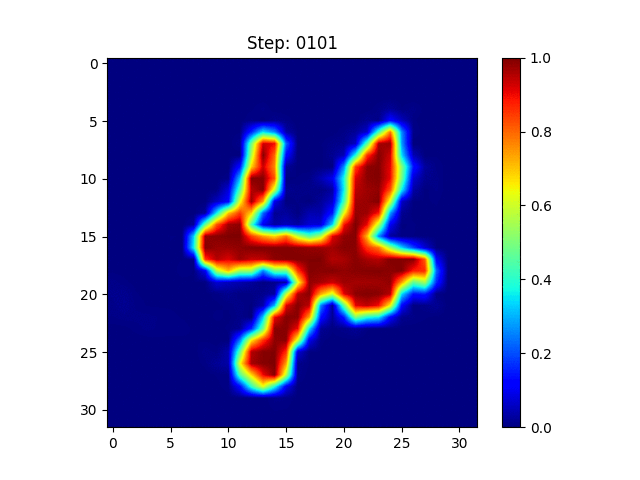}
 
    \caption{\footnotesize Continuous warping of `3' to `4' using our Warping NODE in GAN setup, when the available data $x_0$ and $x_1$ is unpaired.}
    \vspace{-15pt}
    \label{fig:3to4}
\end{figure}

\section{Experiments}
We seek to demonstrate the ability of our model to generate an animation, given two conceptually far apart frames.

\subsection{Proof of concept}
We start the experimental section by introducing Warping Neural ODE as a generative model in a GAN setup. We train our model to warp samples from a distribution in the shape of handwritten digits `3' to digits `4'. Note that there is no paired data in the sense, none of the `4' corresponds to any specific `3'.
Since we can evaluate the solution of ODE in \eqref{eq:diff} at arbitrary times $t$, in Fig.~\ref{fig:3to4}, we visualize the evolution of $\Phi_t$. For the visualisation purpose we show warped frames only for limited time steps, but we provide a corresponding `.gif' file with a smooth animation on the website. 
It is evident from the `conceptual' experiment that our model can indeed learn a smooth and complicated diffeomorphism given only two frames, initial and final states.  However, while the underlying spatial transformation from digits `3' to `4' is complex,
the constant background colors makes it easier to find a real diffeomorphism. Which is changed in the second set of experiments.

\subsection{Facial expressions}
\begin{figure}[!ht]

     \includegraphics[width=0.159\columnwidth, trim={4cm, 1.4cm, 4cm, 2cm}, clip]{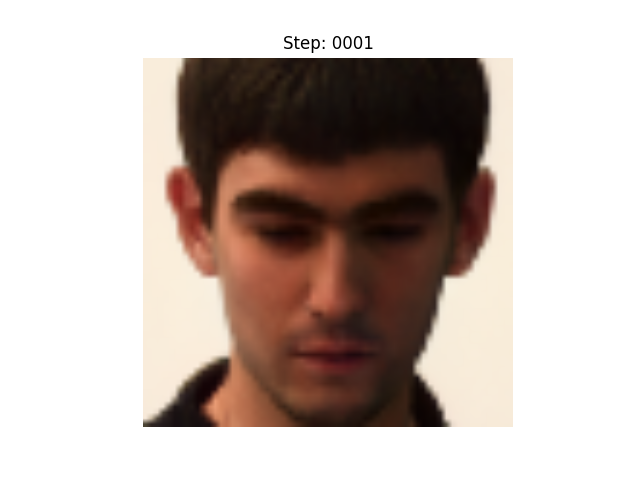}
     \includegraphics[width=0.159\columnwidth, trim={4cm, 1.4cm, 4cm, 2cm}, clip]{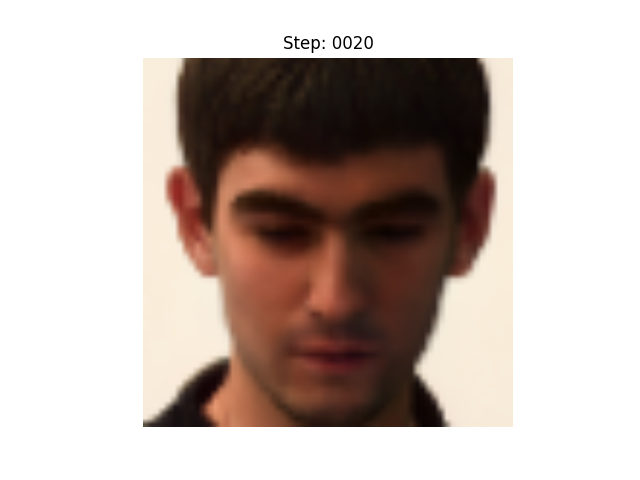}
     \includegraphics[width=0.159\columnwidth, trim={4cm, 1.4cm, 4cm, 2cm}, clip]{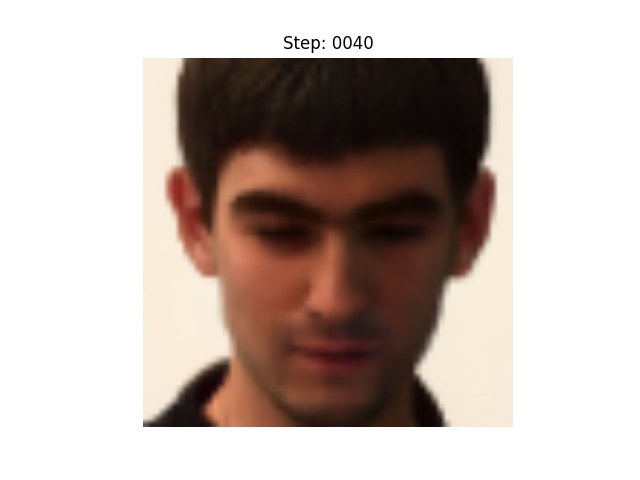}
     \includegraphics[width=0.159\columnwidth, trim={4cm, 1.4cm, 4cm, 2cm}, clip]{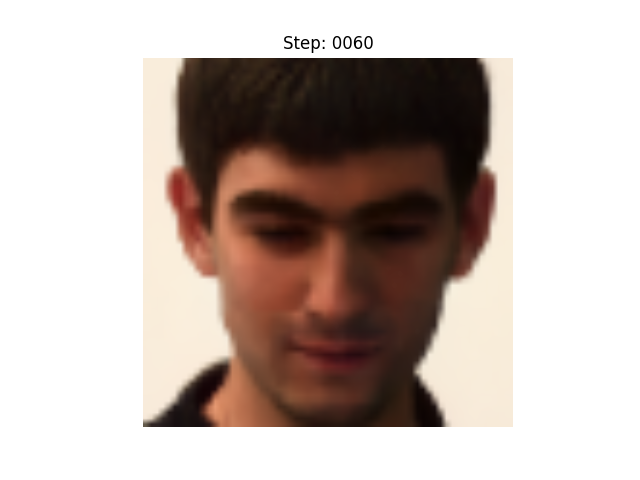}
     \includegraphics[width=0.159\columnwidth, trim={4cm, 1.4cm, 4cm, 2cm}, clip]{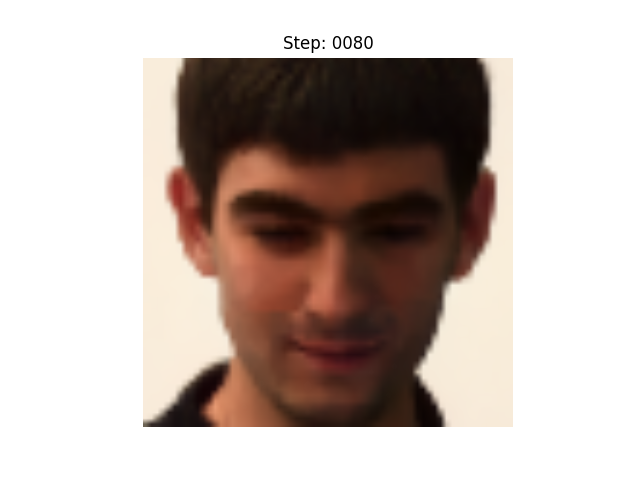}
     \includegraphics[width=0.159\columnwidth, trim={4cm, 1.4cm, 4cm, 2cm}, clip]{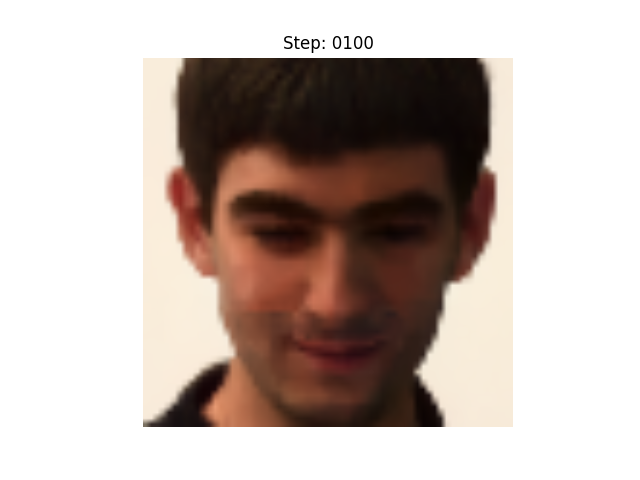}
     
     \includegraphics[width=0.159\columnwidth, trim={4cm, 1.4cm, 4cm, 2cm}, clip]{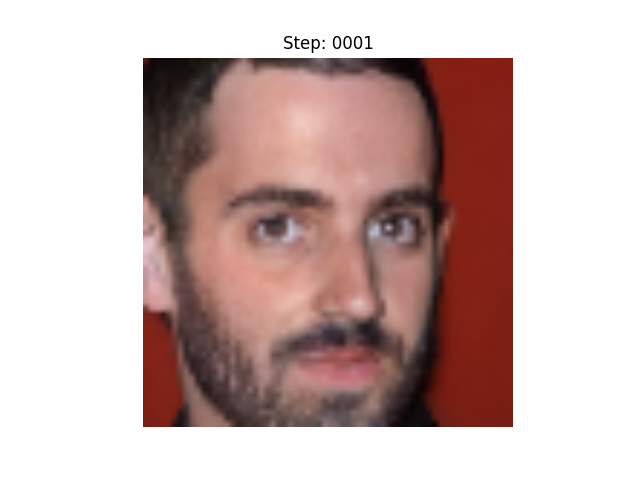}
     \includegraphics[width=0.159\columnwidth, trim={4cm, 1.4cm, 4cm, 2cm}, clip]{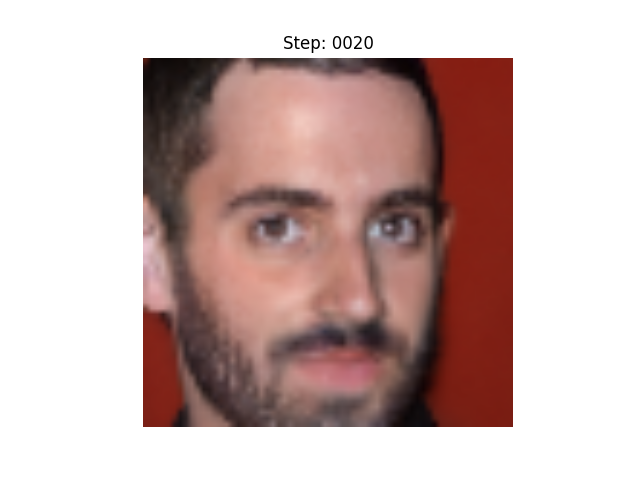}
     \includegraphics[width=0.159\columnwidth, trim={4cm, 1.4cm, 4cm, 2cm}, clip]{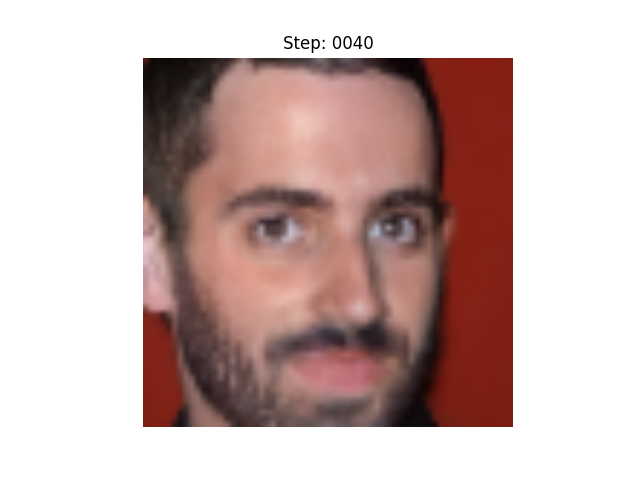}
     \includegraphics[width=0.159\columnwidth, trim={4cm, 1.4cm, 4cm, 2cm}, clip]{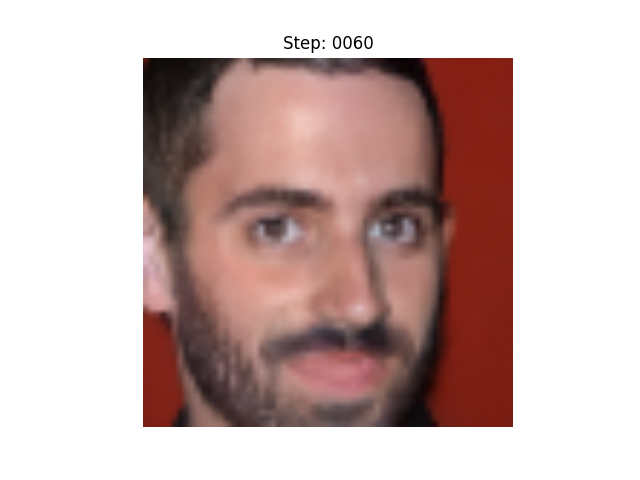}
     \includegraphics[width=0.159\columnwidth, trim={4cm, 1.4cm, 4cm, 2cm}, clip]{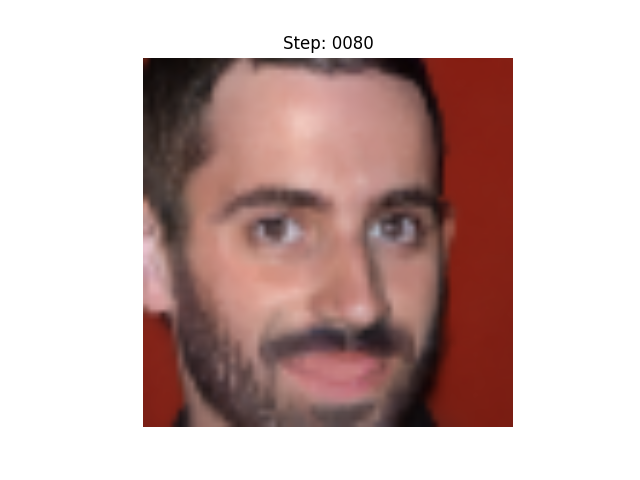}
     \includegraphics[width=0.159\columnwidth, trim={4cm, 1.4cm, 4cm, 2cm}, clip]{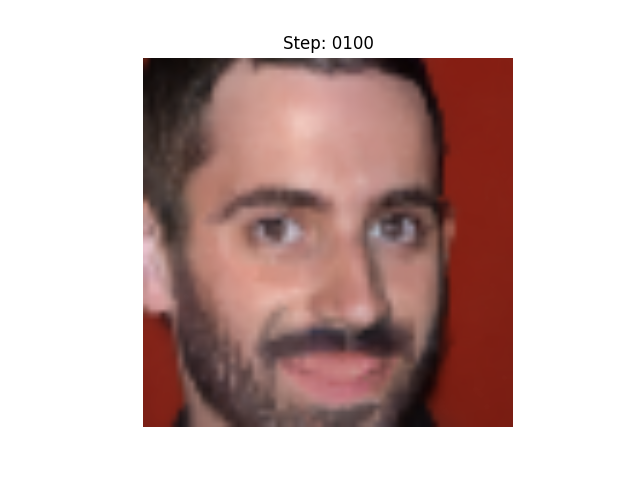}
     
     \includegraphics[width=0.159\columnwidth, trim={4cm, 1.4cm, 4cm, 2cm}, clip]{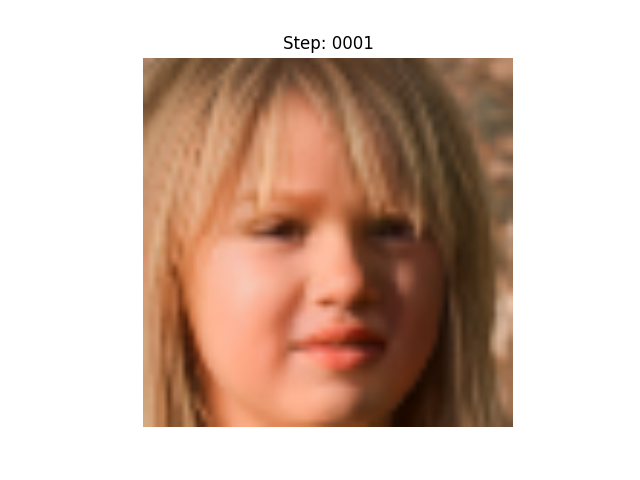}
     \includegraphics[width=0.159\columnwidth, trim={4cm, 1.4cm, 4cm, 2cm}, clip]{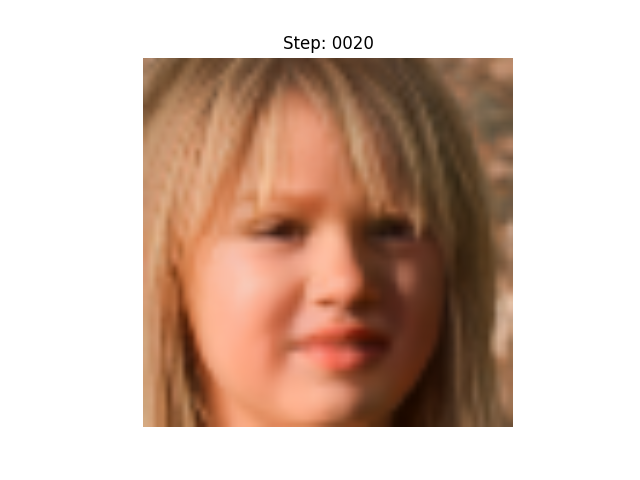}
     \includegraphics[width=0.159\columnwidth, trim={4cm, 1.4cm, 4cm, 2cm}, clip]{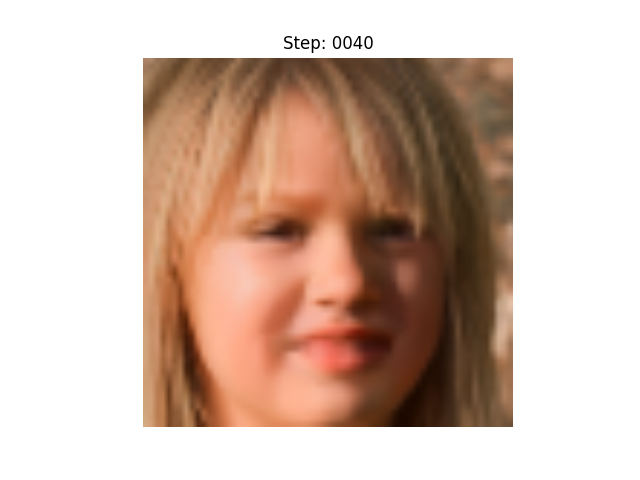}
     \includegraphics[width=0.159\columnwidth, trim={4cm, 1.4cm, 4cm, 2cm}, clip]{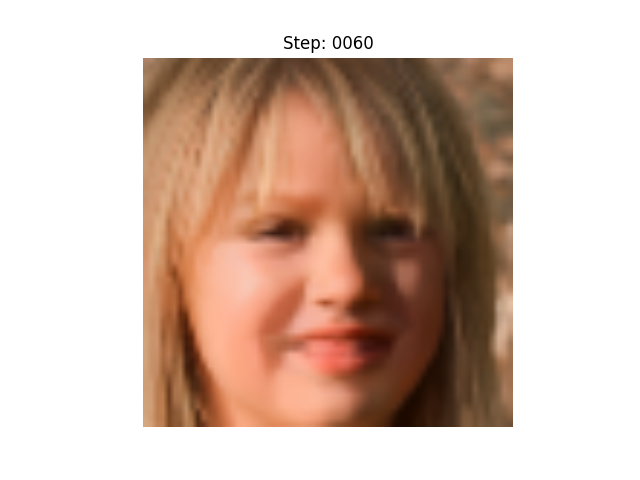}
     \includegraphics[width=0.159\columnwidth, trim={4cm, 1.4cm, 4cm, 2cm}, clip]{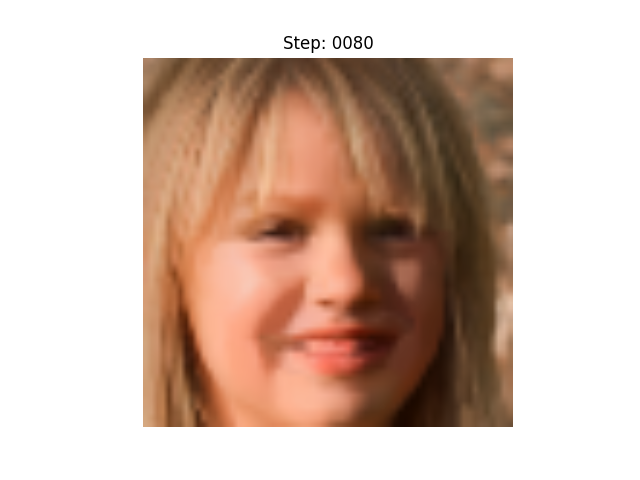}
     \includegraphics[width=0.159\columnwidth, trim={4cm, 1.4cm, 4cm, 2cm}, clip]{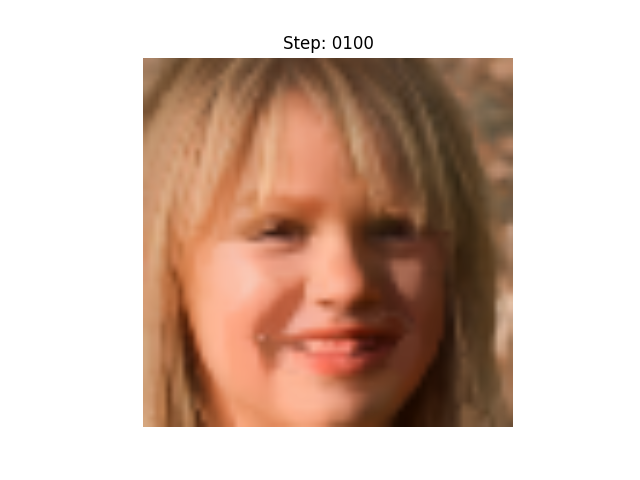}

     \includegraphics[width=0.159\columnwidth, trim={4cm, 1.4cm, 4cm, 2cm}, clip]{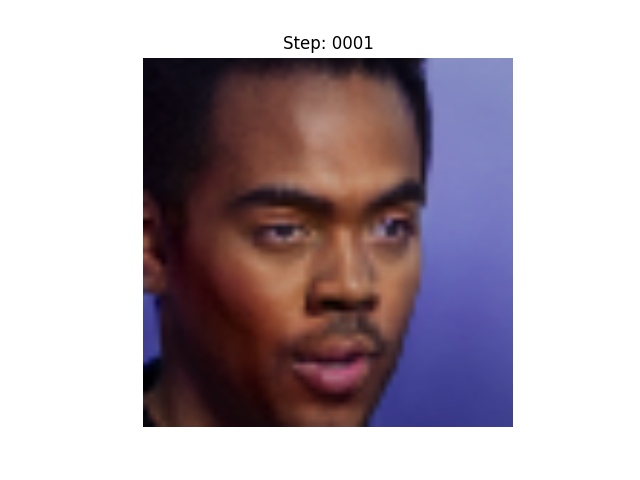}
     \includegraphics[width=0.159\columnwidth, trim={4cm, 1.4cm, 4cm, 2cm}, clip]{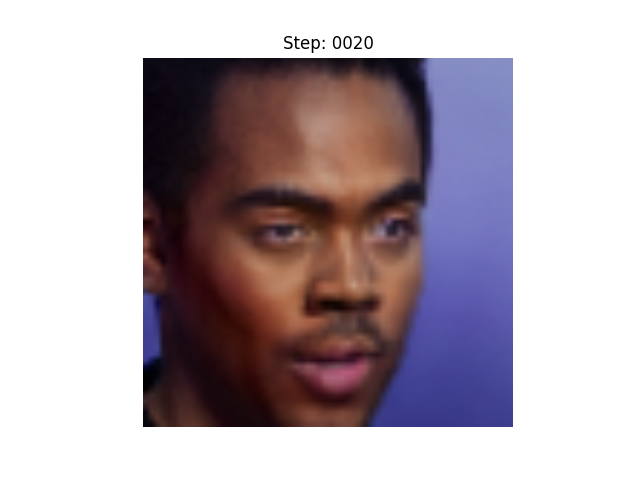}
     \includegraphics[width=0.159\columnwidth, trim={4cm, 1.4cm, 4cm, 2cm}, clip]{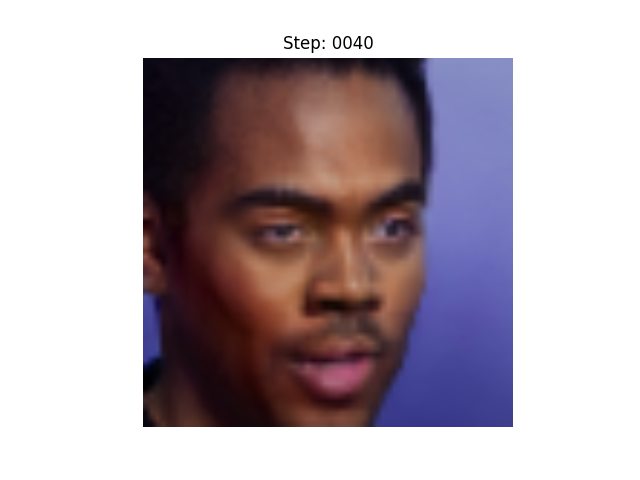}
     \includegraphics[width=0.159\columnwidth, trim={4cm, 1.4cm, 4cm, 2cm}, clip]{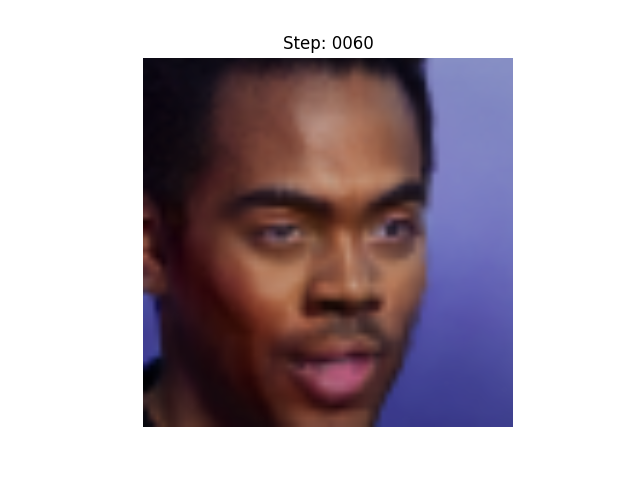}
     \includegraphics[width=0.159\columnwidth, trim={4cm, 1.4cm, 4cm, 2cm}, clip]{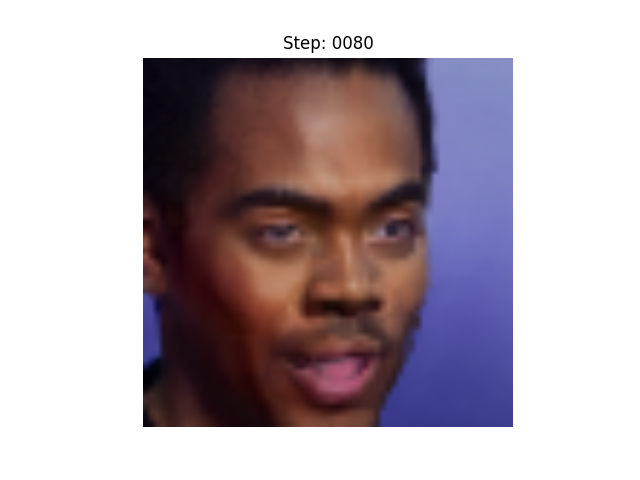}
     \includegraphics[width=0.159\columnwidth, trim={4cm, 1.4cm, 4cm, 2cm}, clip]{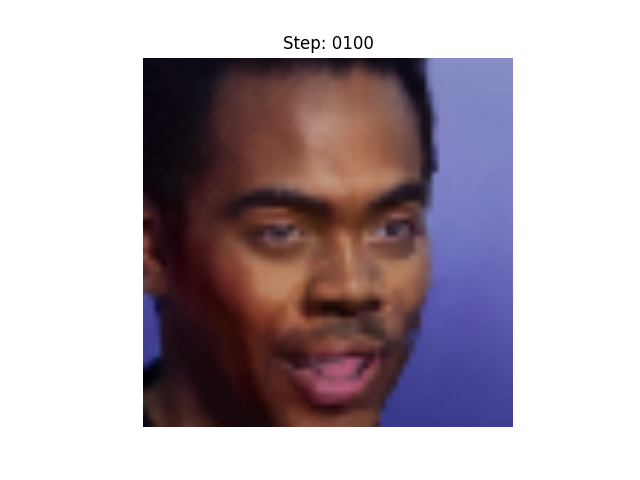}
    \caption{\footnotesize Interpolated frames using $L_2$ loss as paired images are given.}
    \vspace{-10pt}
    \label{fig:l2}
\end{figure}

\begin{figure}[!ht]
     \includegraphics[width=0.159\columnwidth, trim={4cm, 1.4cm, 4cm, 2cm}, clip]{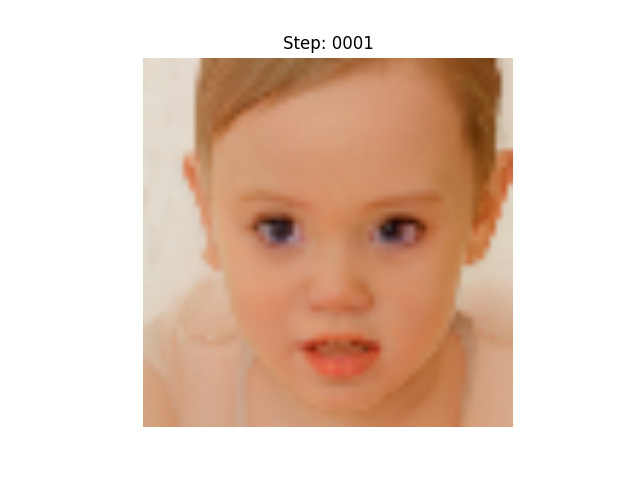}
     \includegraphics[width=0.159\columnwidth, trim={4cm, 1.4cm, 4cm, 2cm}, clip]{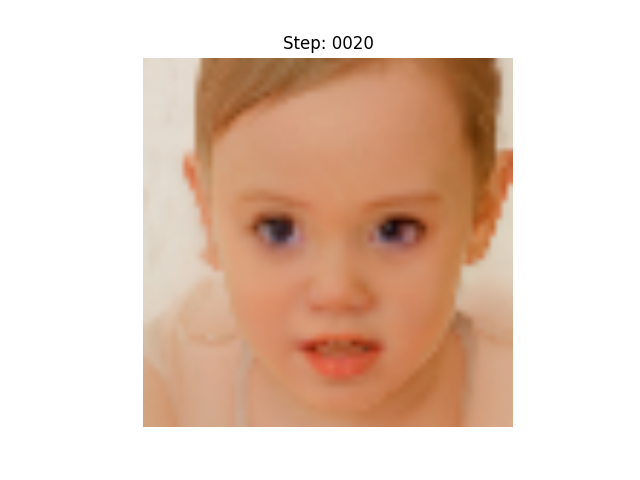}
     \includegraphics[width=0.159\columnwidth, trim={4cm, 1.4cm, 4cm, 2cm}, clip]{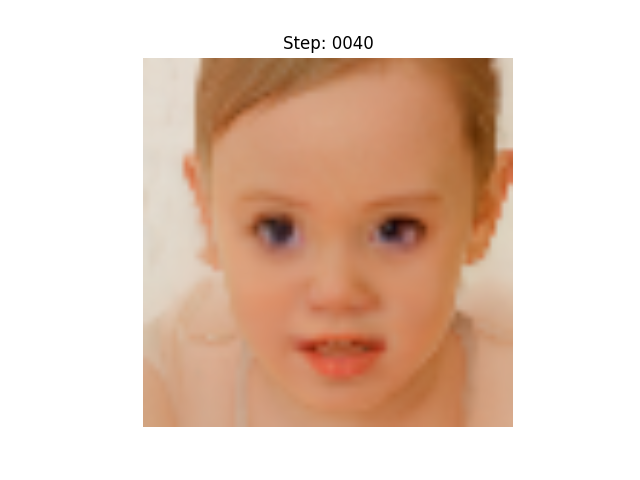}
     \includegraphics[width=0.159\columnwidth, trim={4cm, 1.4cm, 4cm, 2cm}, clip]{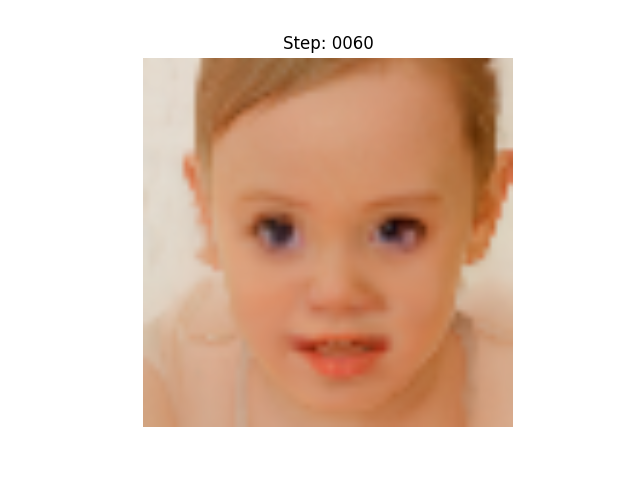}
     \includegraphics[width=0.159\columnwidth, trim={4cm, 1.4cm, 4cm, 2cm}, clip]{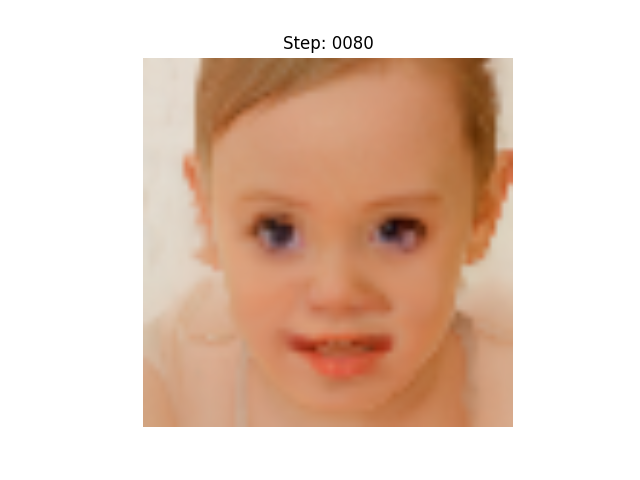}
     \includegraphics[width=0.159\columnwidth, trim={4cm, 1.4cm, 4cm, 2cm}, clip]{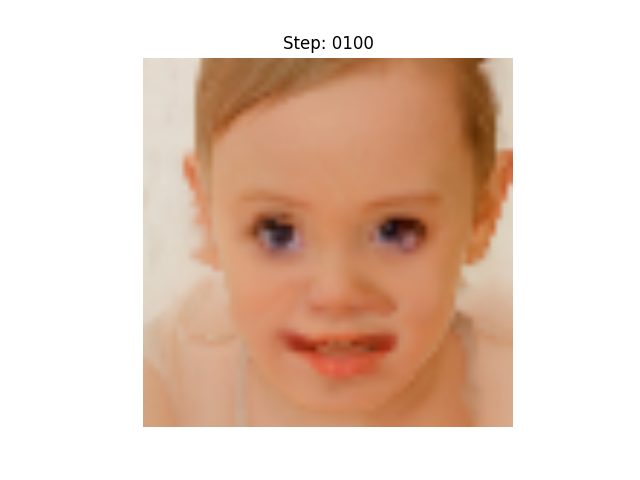}
     
     \includegraphics[width=0.159\columnwidth, trim={4cm, 1.4cm, 4cm, 2cm}, clip]{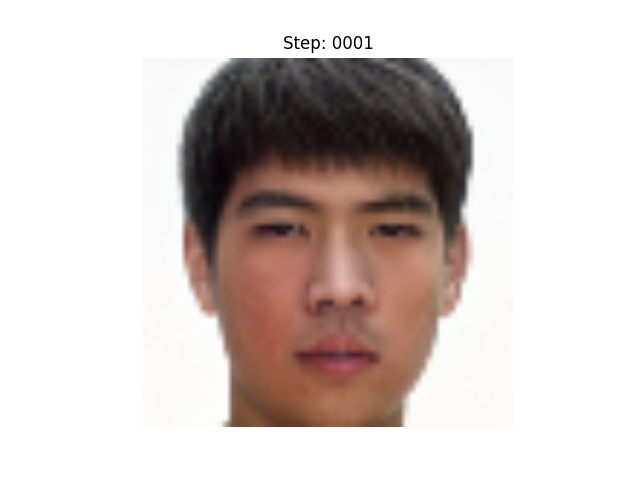}
     \includegraphics[width=0.159\columnwidth, trim={4cm, 1.4cm, 4cm, 2cm}, clip]{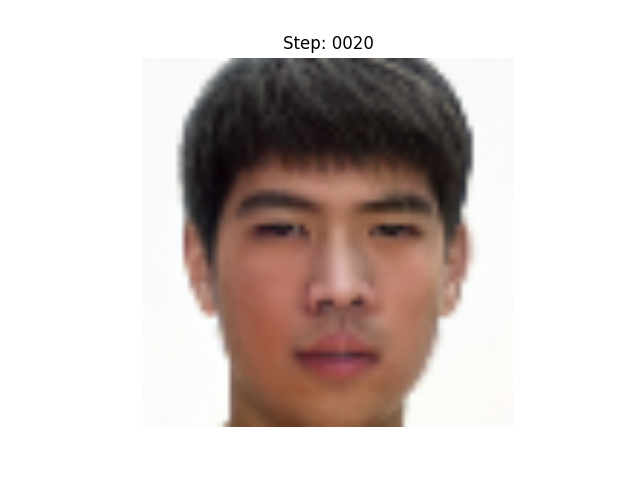}
     \includegraphics[width=0.159\columnwidth, trim={4cm, 1.4cm, 4cm, 2cm}, clip]{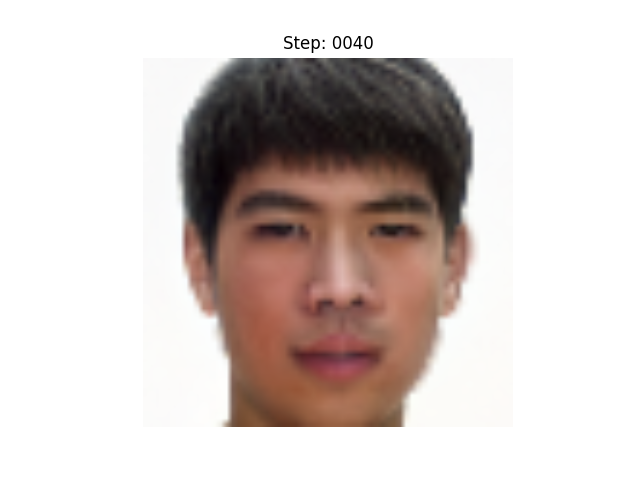}
     \includegraphics[width=0.159\columnwidth, trim={4cm, 1.4cm, 4cm, 2cm}, clip]{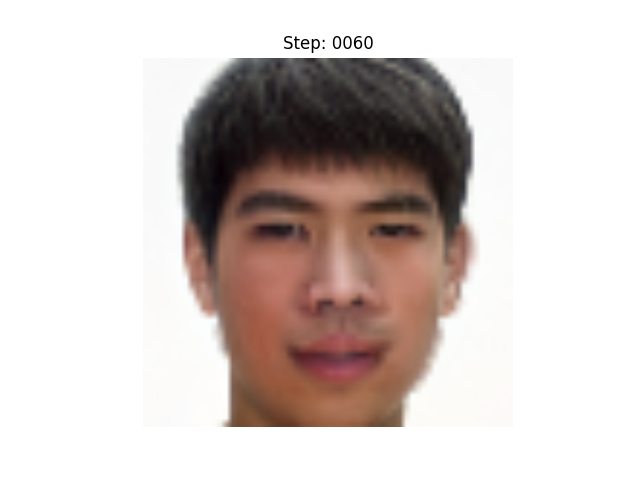}
     \includegraphics[width=0.159\columnwidth, trim={4cm, 1.4cm, 4cm, 2cm}, clip]{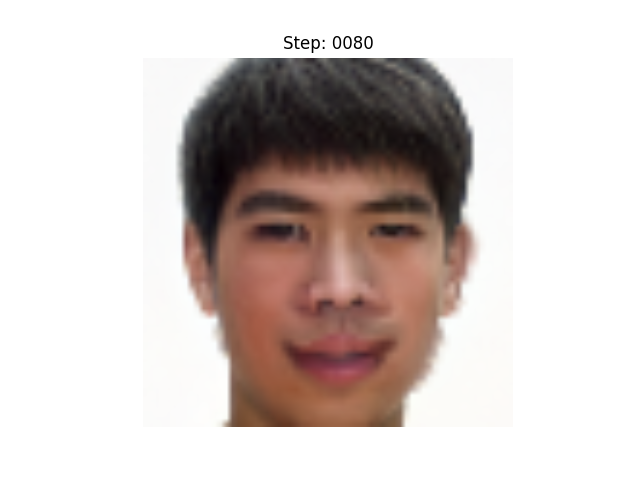}
     \includegraphics[width=0.159\columnwidth, trim={4cm, 1.4cm, 4cm, 2cm}, clip]{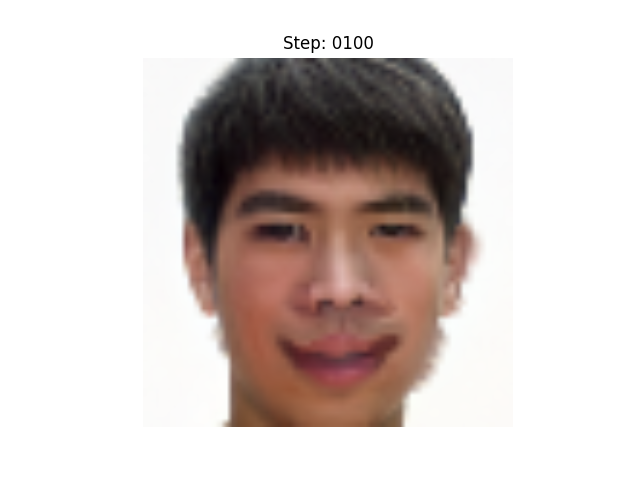}
     
     \includegraphics[width=0.159\columnwidth, trim={4cm, 1.4cm, 4cm, 2cm}, clip]{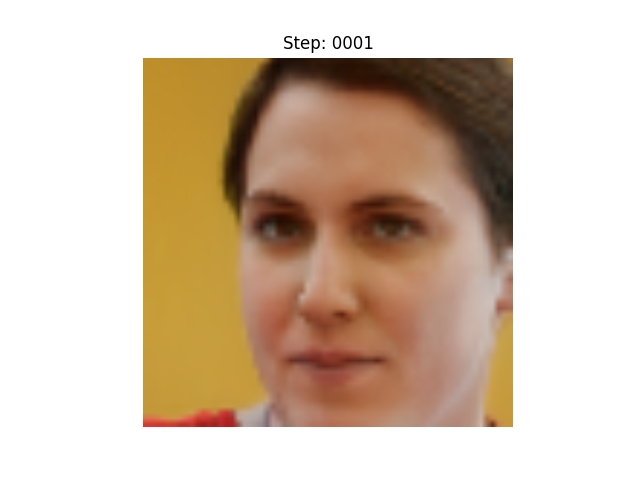}
     \includegraphics[width=0.159\columnwidth, trim={4cm, 1.4cm, 4cm, 2cm}, clip]{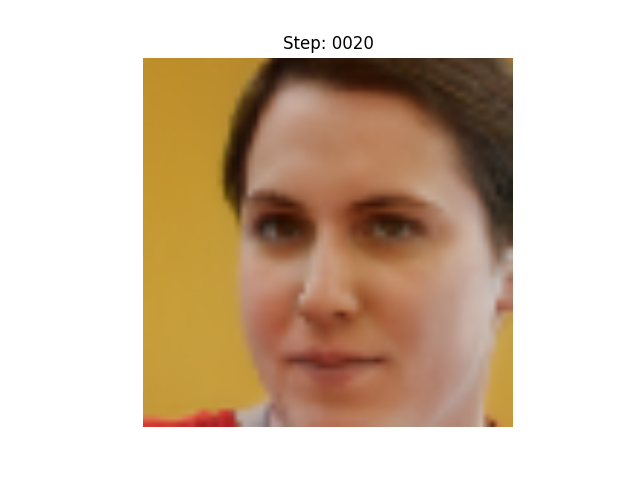}
     \includegraphics[width=0.159\columnwidth, trim={4cm, 1.4cm, 4cm, 2cm}, clip]{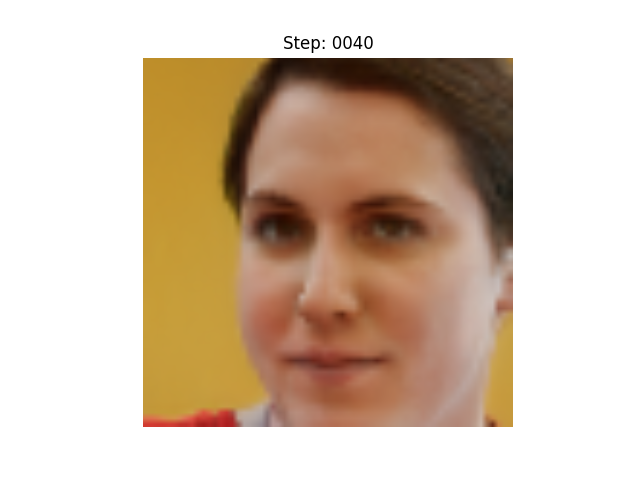}
     \includegraphics[width=0.159\columnwidth, trim={4cm, 1.4cm, 4cm, 2cm}, clip]{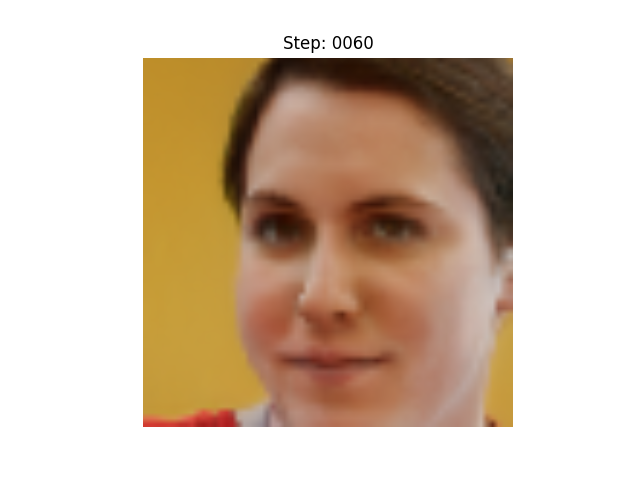}
     \includegraphics[width=0.159\columnwidth, trim={4cm, 1.4cm, 4cm, 2cm}, clip]{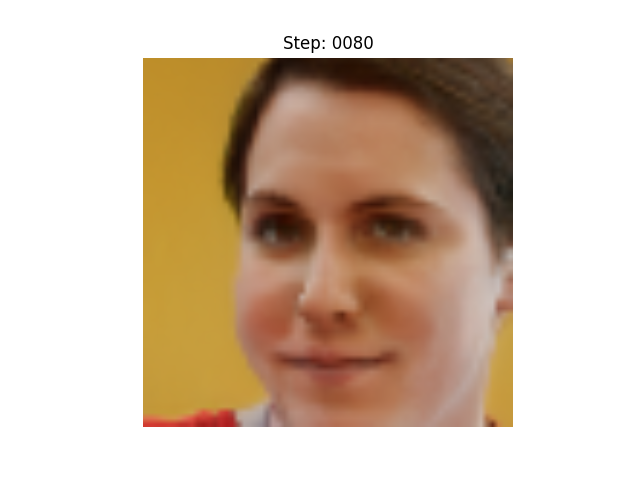}
     \includegraphics[width=0.159\columnwidth, trim={4cm, 1.4cm, 4cm, 2cm}, clip]{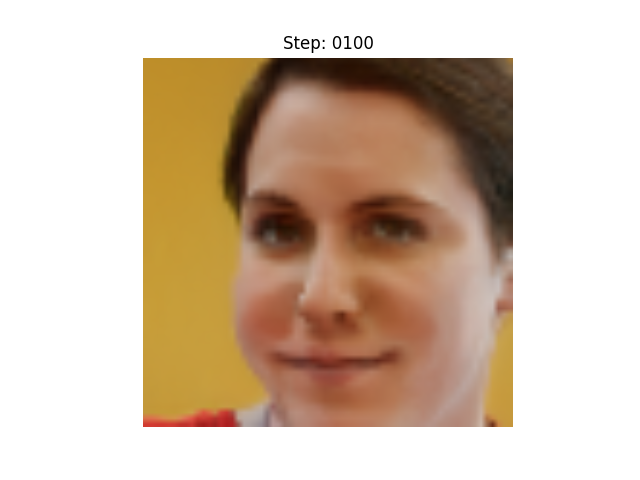}

     \includegraphics[width=0.159\columnwidth, trim={4cm, 1.4cm, 4cm, 2cm}, clip]{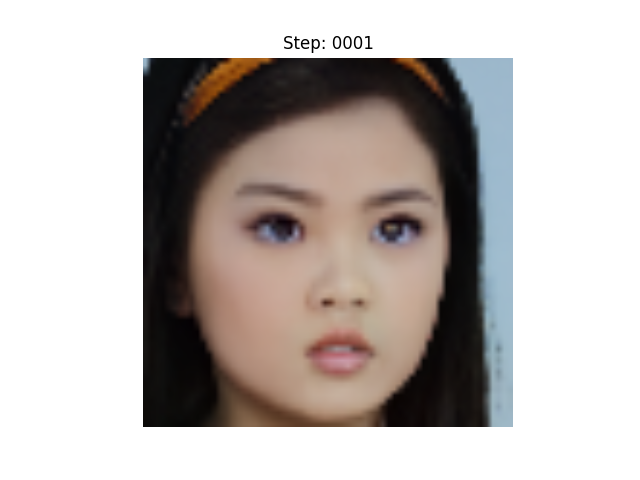}
     \includegraphics[width=0.159\columnwidth, trim={4cm, 1.4cm, 4cm, 2cm}, clip]{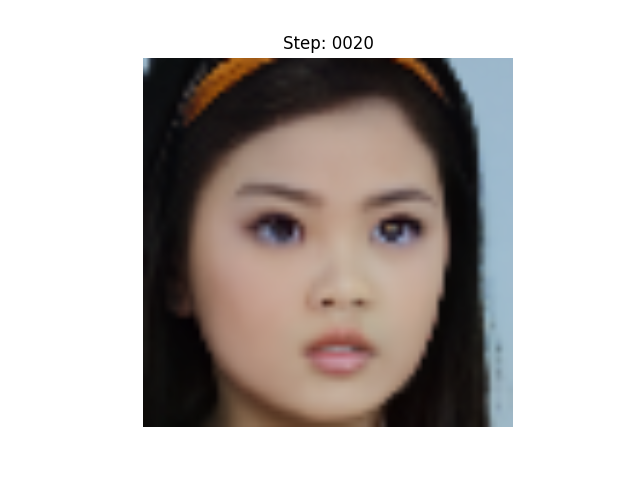}
     \includegraphics[width=0.159\columnwidth, trim={4cm, 1.4cm, 4cm, 2cm}, clip]{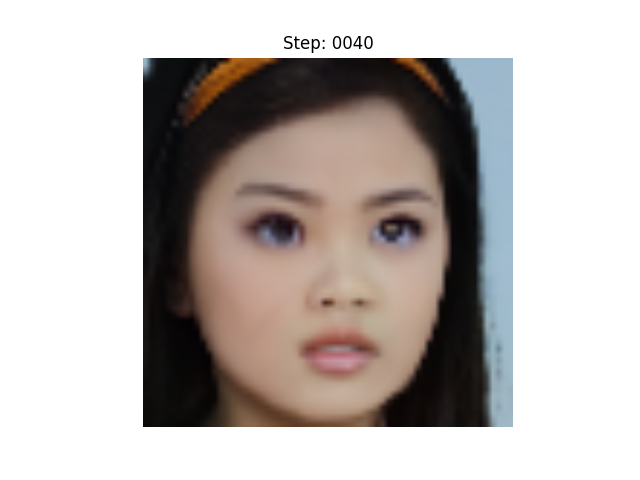}
     \includegraphics[width=0.159\columnwidth, trim={4cm, 1.4cm, 4cm, 2cm}, clip]{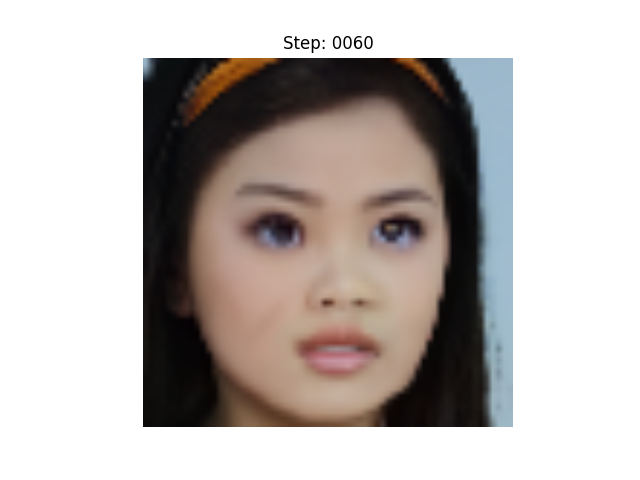}
     \includegraphics[width=0.159\columnwidth, trim={4cm, 1.4cm, 4cm, 2cm}, clip]{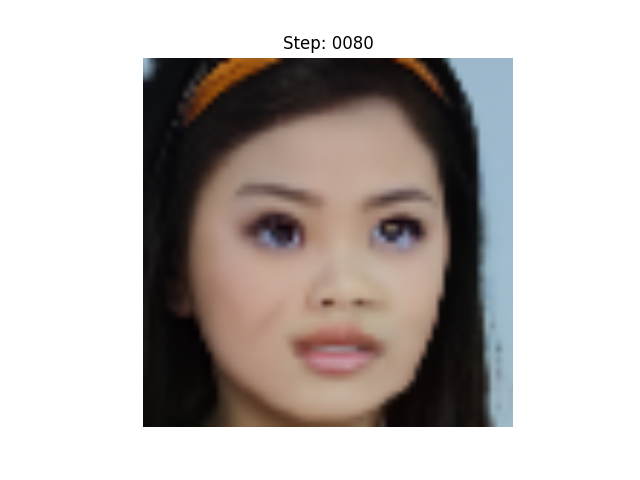}
     \includegraphics[width=0.159\columnwidth, trim={4cm, 1.4cm, 4cm, 2cm}, clip]{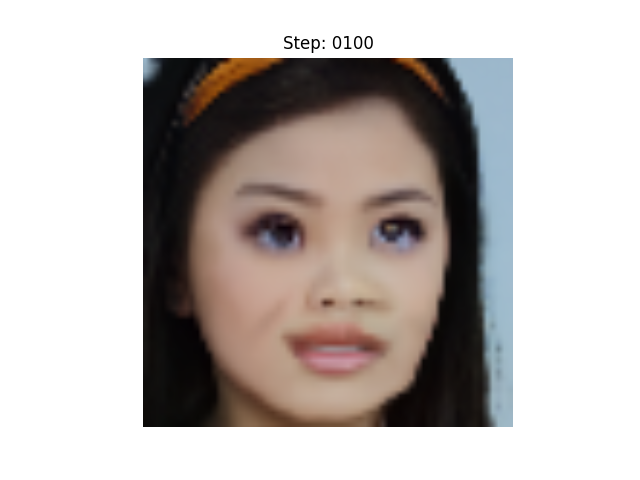}
    \caption{\footnotesize Interpolated frames using GAN setup as paired images are not available.}
    \vspace{-15pt}
    \label{fig:gan}
\end{figure}

Here we evaluate the ability of our model to create an animation of facial expression. Namely, we demonstrate our results on generating the animation, which correspond to emotion `smile'.
As we mentioned earlier, Warping NODE can be applied in different setups: {\bf (a)} when paired images exist, that is, we have 2 frames of the same person, non-smiling and smiling, and thus, $L_2$ reconstruction loss can be applied; {\bf (b)} when paired images do not exist, that is, we have a set of non-smiling faces and smiling-faces, and notion of divergence, similar to GAN have to be used. 
We run experiments in both setups, where Figure~\ref{fig:l2} demonstrates results for $L_2$ training and  Figure~\ref{fig:gan} in a GAN setup. Note that for both experiments, only 2 frames were available for training, namely beginning of the emotion (no smile) and emotion itself (smile), which corresponds to first and last columns in Figures \ref{fig:l2} and \ref{fig:gan}.

While images generated in GAN setup do not demonstrate the same level of quality as in $L_2$ loss, this is expected since no paired images were given to generate warping.

\subsection{Limitations of the method}
Note that to achieve natural animation, our model is based on two important concepts, changes in the images are created by a {\bf (a)} spatial diffeomorphism transformations, that is during the animation all frames contain the same amount of information with respect to the domain (colors),
and {\bf (b)} the continuity of this transformation. 
While there is no problem with continuity {\bf (b)}, the spatial transformation might not be the best fit in every case. For example, if a person turns a head on 180 degrees, 
then this corresponds to a completely new view of a person. That can be challenging to model with just spatial transformations, since we cannot completely disregard color information. 

\section{Conclusions}

This paper provides a new framework, Warping Neural ODE, to generate a smooth animation/VFI between two conceptually far apart frames. 
Experimentally we demonstrated the ability of our model to generate a smooth animation of facial expression, like smile. 
In addition, we show that our framework can be used in both setups, where paired images are available (i.e. the $L_2$ loss can be exploited) and no paired images are available (GAN setup).

We believe that the ability to generate a smooth animation on limited amount of temporal data (namely two time points) can be beneficial to a vision community, including business application in entertainment sphere, by generating new temporal data sets, increasing rate of FPS and other. 

\textbf{Societal impacts.}
We believe that our paper provides an overall beneficial impact on vision community and cannot see an application of it in the harmful way.

{\small
\bibliographystyle{ieee_fullname}
\bibliography{refs}

\begin{thebibliography}{10}\itemsep=-1pt

\bibitem{arjovsky2017wasserstein}
Martin Arjovsky, Soumith Chintala, and L{\'e}on Bottou.
\newblock Wasserstein generative adversarial networks.
\newblock In {\em International conference on machine learning}, pages
  214--223. PMLR, 2017.

\bibitem{ashburner2007fast}
John Ashburner.
\newblock A fast diffeomorphic image registration algorithm.
\newblock {\em Neuroimage}, 38(1):95--113, 2007.

\bibitem{chen2018neural}
Ricky~TQ Chen, Yulia Rubanova, Jesse Bettencourt, and David~K Duvenaud.
\newblock Neural ordinary differential equations.
\newblock {\em Advances in neural information processing systems}, 31, 2018.

\bibitem{fuglede2004jensen}
Bent Fuglede and Flemming Topsoe.
\newblock Jensen-shannon divergence and hilbert space embedding.
\newblock In {\em International Symposium onInformation Theory, 2004. ISIT
  2004. Proceedings.}, page~31. IEEE, 2004.

\bibitem{gulrajani2017improved}
Ishaan Gulrajani, Faruk Ahmed, Martin Arjovsky, Vincent Dumoulin, and Aaron
  Courville.
\newblock Improved training of wasserstein gans.
\newblock {\em arXiv preprint arXiv:1704.00028}, 2017.

\bibitem{hershey2007approximating}
John~R Hershey and Peder~A Olsen.
\newblock Approximating the kullback leibler divergence between gaussian
  mixture models.
\newblock In {\em 2007 IEEE International Conference on Acoustics, Speech and
  Signal Processing-ICASSP'07}, volume~4, pages IV--317. IEEE, 2007.

\bibitem{iserles2000lie}
Arieh Iserles, Hans~Z Munthe-Kaas, Syvert~P N{\o}rsett, and Antonella Zanna.
\newblock Lie-group methods.
\newblock {\em Acta numerica}, 9:215--365, 2000.

\bibitem{jaderberg2015spatial}
Max Jaderberg, Karen Simonyan, Andrew Zisserman, et~al.
\newblock Spatial transformer networks.
\newblock {\em Advances in neural information processing systems},
  28:2017--2025, 2015.

\bibitem{jiang2018super}
Huaizu Jiang, Deqing Sun, Varun Jampani, Ming-Hsuan Yang, Erik Learned-Miller,
  and Jan Kautz.
\newblock Super slomo: High quality estimation of multiple intermediate frames
  for video interpolation.
\newblock In {\em Proceedings of the IEEE conference on computer vision and
  pattern recognition}, pages 9000--9008, 2018.

\bibitem{karras2019style}
Tero Karras, Samuli Laine, and Timo Aila.
\newblock A style-based generator architecture for generative adversarial
  networks.
\newblock In {\em Proceedings of the IEEE/CVF conference on computer vision and
  pattern recognition}, pages 4401--4410, 2019.

\bibitem{kuang2019cycle}
Dongyang Kuang.
\newblock Cycle-consistent training for reducing negative jacobian determinant
  in deep registration networks.
\newblock In {\em International Workshop on Simulation and Synthesis in Medical
  Imaging}, pages 120--129. Springer, 2019.

\bibitem{mueggler2017event}
Elias Mueggler, Henri Rebecq, Guillermo Gallego, Tobi Delbruck, and Davide
  Scaramuzza.
\newblock The event-camera dataset and simulator: Event-based data for pose
  estimation, visual odometry, and slam.
\newblock {\em The International Journal of Robotics Research}, 36(2):142--149,
  2017.

\bibitem{nazarovs2022rf}
Jurijs Nazarovs, Zhichun Huang, Songwong Tasneeyapant, Rudrasis Chakraborty,
  and Vikas Singh.
\newblock Understanding uncertainty maps in vision with statistical testing.
\newblock In {\em Proceedings of the IEEE/CVF Conference on Computer Vision and
  Pattern Recognition}, 2022.

\bibitem{ohrnell2020lie}
Carl {\"O}hrnell.
\newblock Lie groups and pde, 2020.

\bibitem{rousseau2020residual}
Fran{\c{c}}ois Rousseau, Lucas Drumetz, and Ronan Fablet.
\newblock Residual networks as flows of diffeomorphisms.
\newblock {\em Journal of Mathematical Imaging and Vision}, 62(3):365--375,
  2020.

\bibitem{ruschendorf1985wasserstein}
Ludger R{\"u}schendorf.
\newblock The wasserstein distance and approximation theorems.
\newblock {\em Probability Theory and Related Fields}, 70(1):117--129, 1985.

\end{thebibliography}
}

\end{document}